\documentclass[letterpaper, 10 pt, conference]{ieeeconf}

\IEEEoverridecommandlockouts
\usepackage{afterpage}
\usepackage{xspace}
\usepackage{amssymb,paralist,epsfig,standalone,bm,placeins}  % assumes amsmath package installed
\usepackage{graphicx} % for pdf, bitmapped graphics files
\usepackage{xcolor}
\usepackage{multirow}
\usepackage{makecell}
\usepackage{multicol}
\usepackage[utf8]{inputenc}
\usepackage{amsmath,amsfonts,booktabs,cite} % general useful stuff
\usepackage{siunitx,textcomp} % for the degree symbol
\usepackage[hidelinks]{hyperref} % to have hyperlinks and for the \ref macros
\let\labelindent\relax
\usepackage[inline]{enumitem} % inline enumerate
\usepackage[nolist]{acronym} % acronyms by the \ac{label} command
\usepackage{caption}
\usepackage{tikz,standalone,pgfplots}
\usetikzlibrary{patterns}
\usepackage{pgfplots, pgfplotstable}
\graphicspath{{figures_tex/}}

\usepackage{changes}
\usepackage[ruled,vlined]{algorithm2e}

\usepackage{subcaption}
\usepackage{fp}
\usepackage{forloop}
\usepackage{siunitx} %for printing
\usepackage{contour}
\usepackage{color}
\usepackage{booktabs} % for professional tables
\usepackage{multirow} % For merging multiple rows
\usepackage{mathtools} %for multlined

\usepackage[many]{tcolorbox}
\usepackage{minted}
\usepackage{pgf-pie}

\makeatletter
\newcommand{\removelatexerror}{\let\@latex@error\@gobble}
\makeatother
\pgfplotsset{compat=1.12}
\newacro{llm}[LLM]{large language model}
\newacro{vlm}[VLM]{vision language model}
\newacro{vla}[VLA]{vision-language-action model}
\newacro{methodname}[IA-VLA]{Input Augmentation for Vision-Language-Action models}
\newcommand{\figref}[1]{\hyperref[#1]{Fig.~\ref*{#1}}}
\newcommand{\tabref}[1]{\hyperref[#1]{Table~\ref*{#1}}}
\newcommand{\secref}[1]{\hyperref[#1]{Section~\ref*{#1}}}
\newcommand{\algoref}[1]{\hyperref[#1]{Algorithm~\ref*{#1}}}

\def\eg{, \textit{e.g.}, }

\definecolor{findOptimalPartition}{HTML}{D7191C}
\definecolor{storeClusterComponent}{HTML}{FDAE61}
\definecolor{dbscan}{HTML}{ABDDA4}
\definecolor{constructCluster}{HTML}{2B83BA}

\title{\LARGE \bf
IA-VLA: Input Augmentation for Vision-Language-Action models in settings with semantically complex tasks
}

\author{Eric~Hannus$^{1}$, Miika~Malin$^{2}$, Tran~Nguyen~Le$^{3}$, Ville~Kyrki$^{1}$%
\thanks{This work was financially supported by the Research Council of Finland project 345661, and the Business Finland decision 9249/31/2021. We acknowledge the computational resources provided by the Aalto Science-IT project and CSC – IT Center for Science, Finland. We also acknowledge the use of MIDAS research infrastructure of Aalto School of Electrical Engineering.}
\thanks{$^{1}$ Intelligent Robotics Group at the Department of Electrical Engineering and
Automation, School of Electrical Engineering, Aalto University, Espoo, Finland.
\texttt{\{firstname.lastname\}{@}aalto.fi}}
\thanks{$^{2}$ Biomimetics and Intelligent Systems Group at the Faculty of Information Technology and Electrical Engineering, University of Oulu, Oulu, Finland. \texttt{\{firstname.lastname\}{@}oulu.fi}}
\thanks{$^{3}$ Section of Mechanical Technology at the Department of Engineering Technology and Didactics, Technical University of Denmark, Denmark. \texttt{tngle@dtu.dk}}
}

\makeatletter
\let\@oldmaketitle\@maketitle% Store \@maketitle
\renewcommand{\@maketitle}{\@oldmaketitle% Update \@maketitle to insert...
  \setcounter{figure}{0}
    \vspace{1.5cm}
    \centering
    \def\svgwidth{\linewidth}
    \fontsize{6}{6}%\selectfont\sf
  \includegraphics[scale=0.23]{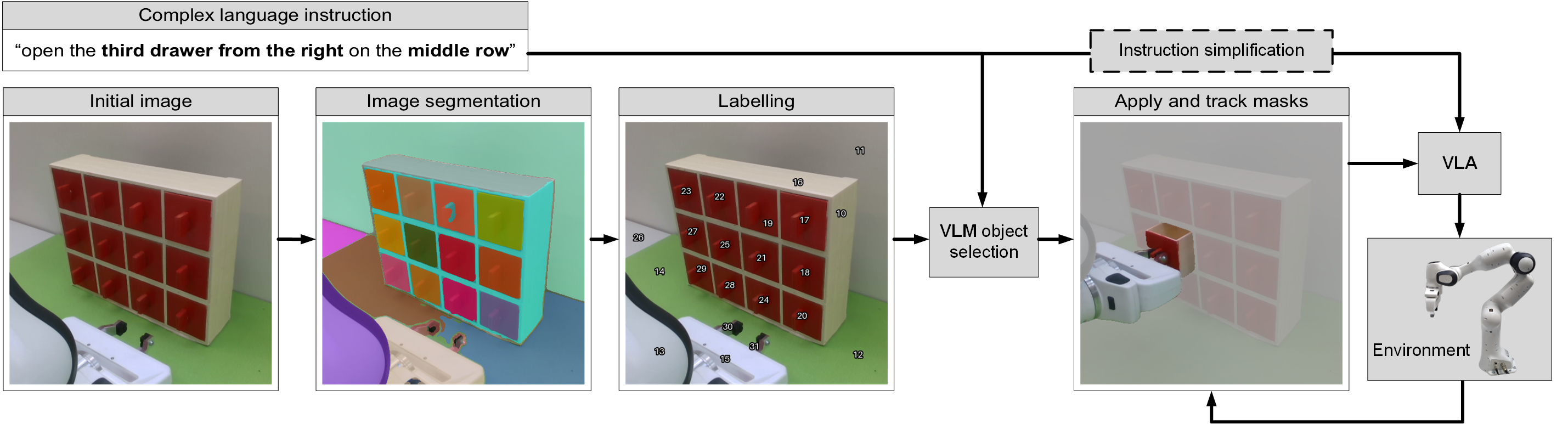}
  \captionof{figure}{
  Semantically complex language instructions, such as those involving the relative positions of objects, pose a difficult challenge for vision-language-action models (VLAs). To address this problem, we propose \acs{methodname}, a framework for augmenting the input to VLAs, that offloads the semantic understanding to a larger vision language model (VLM) with greater semantic understanding. We use semantic segmentation to label image regions which a VLM then uses to identify the masks of the task-relevant objects. The task-relevant objects are highlighted in the VLA input, together with the language instruction which can optionally be simplified. 
 \label{fig:proposed_method}
}}
\makeatother

\begin{document}
\maketitle
\thispagestyle{empty}
\pagestyle{empty}

%%%%%%%%%%%%%%%%%%%%%%%%%%%%%%%%%%%%%%%%%%%%%%%%%%%%%%%%%%%%%%%%%%%%%%%%%%%%%%%%

\begin{abstract}
Vision-language-action models (VLAs) have become an increasingly popular approach for addressing robot manipulation problems in recent years. However, such models need to output actions at a rate suitable for robot control, which limits the size of the language model they can be based on, and consequently, their language understanding capabilities. Manipulation tasks may require complex language instructions, such as identifying target objects by their relative positions, to specify human intention.
Therefore, we introduce \textit{\acs{methodname}}, a framework that utilizes the extensive language understanding of a large vision language model as a pre-processing stage to generate improved context to augment the input of a VLA.
We evaluate the framework on a set of semantically complex tasks which have been underexplored in VLA literature, namely tasks involving visual duplicates, \textit{i.e.}, visually indistinguishable objects. A dataset of three types of scenes with duplicate objects is used to compare a baseline VLA against two augmented variants. The experiments show that the VLA benefits from the augmentation scheme, especially when faced with language instructions that require the VLA to extrapolate from concepts it has seen in the demonstrations. For the code, dataset, and videos, see \url{https://sites.google.com/view/ia-vla}.

\end{abstract}

%%%%%%%% Start of document

\section{Introduction}
\label{sec:introduction}

Operation in varied environments is an essential research problem in robotics.
Ideally, the robots' skills need to generalize to objects and object configurations which the robots have not encountered during training. Building on the advances in deep generative models, multiple generalist robot policies have been proposed in recent years \cite{jang2022BC_Z, brohan2023rt1, zitkovich2023rt2, Ghosh2024_Octo, kim2024_openvla, li2024cogactfoundationalvisionlanguageactionmodel, black2024pi0visionlanguageactionflowmodel, wen2025TinyVLA}. Particularly \acp{vla} \cite{zitkovich2023rt2, kim2024_openvla, li2024cogactfoundationalvisionlanguageactionmodel, black2024pi0visionlanguageactionflowmodel, wen2025TinyVLA}, models that adapt \acp{vlm} to output robot actions, have shown impressive results.

The generalization capabilities of \acp{vlm} are often evaluated with respect to the robustness to variations in backgrounds, lighting conditions, object appearances, and object locations \cite{zitkovich2023rt2, kim2024_openvla, wen2025TinyVLA}. However, from a language-understanding perspective, the model's ability to deal with previously unseen scenes and objects, and to choose the relevant objects among distractors, is especially relevant. Previous works have investigated unseen colors, shapes, or object instances for seen \cite{wen2025TinyVLA, kim2024_openvla, li2024cogactfoundationalvisionlanguageactionmodel} and unseen \cite{zitkovich2023rt2, kim2024_openvla} object categories necessitating general knowledge from the underlying \ac{vlm}. However, complex scenes with multiple visually indistinguishable objects, have not been investigated in \ac{vla} literature. Such scenes require that the object of interest is specified in complicated spatial terms in the language instruction.

We hypothesize that current \ac{vla} may not be equipped with sufficient language understanding to fully interpret such tasks when compared to state-of-the-art \acp{vlm}, as \ac{vla} inference latency scales with model size \cite{wen2025TinyVLA}, and they must be small enough to generate actions at a suitable rate for robot control. Even though there have been works increasing inference speed with diffusion-based approaches \cite{wen2025TinyVLA, li2024cogactfoundationalvisionlanguageactionmodel, black2024pi0visionlanguageactionflowmodel} or parallel decoding \cite{kim2025OpenVLA-OFT} the models are still based on \acp{llm} with under 10 billion parameters.

To tackle these complex tasks, we propose the \textit{\acs{methodname}: \aclu{methodname}} framework presented in Fig. \ref{fig:proposed_method}. The semantically complex input is first processed with a general-purpose \ac{vlm} to identify the object instances relevant to the task using numeric labels added to the image, similar to \cite{yang2023set_of_marks}. Then, the images are augmented to highlight the relevant object instances using their segmentation masks. The segmentation masks, and consequently the highlights, are propagated to future time-steps using a model with mask-tracking capabilities.
As the \ac{vlm} is only used for processing the initial image, we can use a much larger \ac{vlm} than those used by \acp{vla}, which need to process every timestep in a timely fashion. In this framework the \ac{vla}'s role is generating actions for visuomotor control based on the augmented images, and its responsibility for parsing semantics is diminished. Our contributions include:
\begin{itemize}
  \item A framework for augmenting the input of \acp{vla} to better handle tasks specified by semantically complex instructions.
  \item A formulation of visual duplicates, a class of complex tasks previously understudied in the \ac{vla} literature, and an associated dataset.
  \item A concrete implementation of the framework with thorough experimental evaluations in settings with duplicate objects, with 1290 evaluation runs in total.
\end{itemize}

\section{Related work}
\label{sec:related_work}
\subsection{Vision-Language-Action Models}
\label{sec: BackgroundVLA}

Generalist robot policies based on deep generative models have only seen wide-scale research interest in recent years, but there are some trends that can be observed.
Early works used custom architectures and datasets consisting primarily of new demonstration data \cite{reed2022gato, jang2022BC_Z, brohan2023rt1}. Compared to these early models, there have been two major changes.
First, RT-2 \cite{zitkovich2023rt2} introduced the concept of \acp{vla}, models based on \acp{vlm} for generating robot actions, which has since become a popular approach \cite{kim2024_openvla, li2024cogactfoundationalvisionlanguageactionmodel, wen2025TinyVLA, black2024pi0visionlanguageactionflowmodel, kim2025OpenVLA-OFT}. Second, the Open-X-Embodiment (OXE) \cite{ONeill_2024_openxembodiment} collaboration gathered a large-scale repository combining datasets from previous research, and showed that generalist robot policies can benefit from training on data from dissimilar embodiments. Subsets of OXE have since been used in many works\eg \cite{Ghosh2024_Octo, kim2024_openvla, black2024pi0visionlanguageactionflowmodel}. 

The objects of interest are typically uniquely identifiable by their visual features in \ac{vla} research problems, with a few exceptions \cite{zitkovich2023rt2, wen2024oci}.
Semantically difficult tasks such as \textit{"move apple to cup with same color"} or \textit{"pick the object that is different from all other objects"} are studied in \cite{zitkovich2023rt2}, but the object of interest is typically visually distinct. Among the examples presented in the paper, there are some exceptions in which there are duplicate fruits in the scene, like \textit{"put the strawberry into the correct bowl"} and \textit{"place orange in the matching bowl"}. However, in both cases the fruit instance to be manipulated is placed separately, while the remaining fruits of the same category are in a bowl. Consequently, there is no need to interpret textual instructions with even more difficult semantics to uniquely identify the relevant object instance. 
In \cite{wen2024oci} some tasks with duplicates are included, but they are limited to simple positional relationships, where the target object can be identified by referring to it as the \textit{left} or \textit{middle} object. Therefore, complex language instructions involving duplicate objects remain an underexplored problem. 

\subsection{Input Augmentation}
\label{sec: BackgroundAugmentation}

Input augmentation has previously been used to enhance performance of robot policies trained on vision-language demonstrations, usually in scenarios with unseen objects \cite{miyashita2023rosoimprovingroboticpolicy, stone2023openworld, yang2025transferringfoundationmodelsgeneralizable, huang2025ottervisionlanguageactionmodeltextaware, li2025languageguidedobjectcentricdiffusionpolicy, Huang_2025_ROBOGROUND, jia2025qquivariant} and distractor objects or backgrounds \cite{miyashita2023rosoimprovingroboticpolicy, stone2023openworld, hancock2024byovla, yang2025transferringfoundationmodelsgeneralizable, huang2025ottervisionlanguageactionmodeltextaware, li2025languageguidedobjectcentricdiffusionpolicy, Huang_2025_ROBOGROUND}. Some works achieve robustness to changes in camera view and collision avoidance \cite{li2025languageguidedobjectcentricdiffusionpolicy}, address simple spatial relationships \cite{wen2024oci, Huang_2025_ROBOGROUND}, or target specialized tasks like tool handover during surgery \cite{li2024robonursevlaroboticscrubnurse}. Most commonly these augment the vision-language input before the corresponding features are generated, but \cite{huang2025ottervisionlanguageactionmodeltextaware} is an exception that extracts the visual features relevant to the task at each timestep before they are input to the action-generating transformer.
Some works annotate the task-relevant objects with bounding boxes \cite{wen2024oci}, single-pixel masks \cite{stone2023openworld}, segmentation masks \cite{yang2025transferringfoundationmodelsgeneralizable, li2025languageguidedobjectcentricdiffusionpolicy, Huang_2025_ROBOGROUND} or pixel-level value maps \cite{jia2025qquivariant} while others use inpainting to replace unseen objects with seen ones \cite{miyashita2023rosoimprovingroboticpolicy} or hide regions with distracting objects or background \cite{hancock2024byovla}.

The most related work is BYOVLA \cite{hancock2024byovla}, which also modifies visual inputs but focuses on mitigating distractors by inpainting irrelevant objects to improve VLA performance. Our work, by contrast, targets a different challenge: semantically complex scenes where the non-target duplicates are not irrelevant noise but critical cues for spatial reasoning and target identification. In addition, BYOVLA is designed purely as a run-time intervention scheme, which augments inputs only during inference. As there is no re-training on the augmented images they aim to retain maximum visual similarity to the pre-training data by only inpainting regions that the VLA is sensitive to. In contrast, we take a different view: practical VLA deployments often require finetuning as environments from the pre-training stage cannot always be reconstructed or new tasks need to be added.
Therefore, we augment images during both training and inference, enabling more substantial visual alterations. Finally, inpainting essentially pretends that objects do not exist, which would make the original instruction uninterpretable in the cases we consider.
Thus, we instead apply a semi-transparent mask for highlighting as we do not need to maximize the visual similarity to the pretraining dataset.

\section{Method}
\label{sec:method}
We study manipulation tasks defined by complex language instructions. In particular, we focus on tasks involving duplicate objects, which necessitate that the target object is specified via spatial relationships.
By duplicate objects we refer to object instances belonging to the same object category, which cannot be identified by their visual features (\textit{e.g.,} color, size, or texture) alone, so that the target objects must be identified by their spatial relationships in the scene.

The framework we propose for dealing with complex language instructions, \ac{methodname}, is shown in Figure \ref{fig:proposed_method}. To identify the relevant object instances, the full input image is segmented using Semantic-SAM \cite{li2023semantic}, numeric tags are added for each mask, and a \ac{vlm} is prompted to select the numbers corresponding to object instances relevant to the task, in a strategy similar to to \cite{yang2023set_of_marks}.

\begin{figure}[hb!]
\centering
\includegraphics[width=0.4\textwidth]{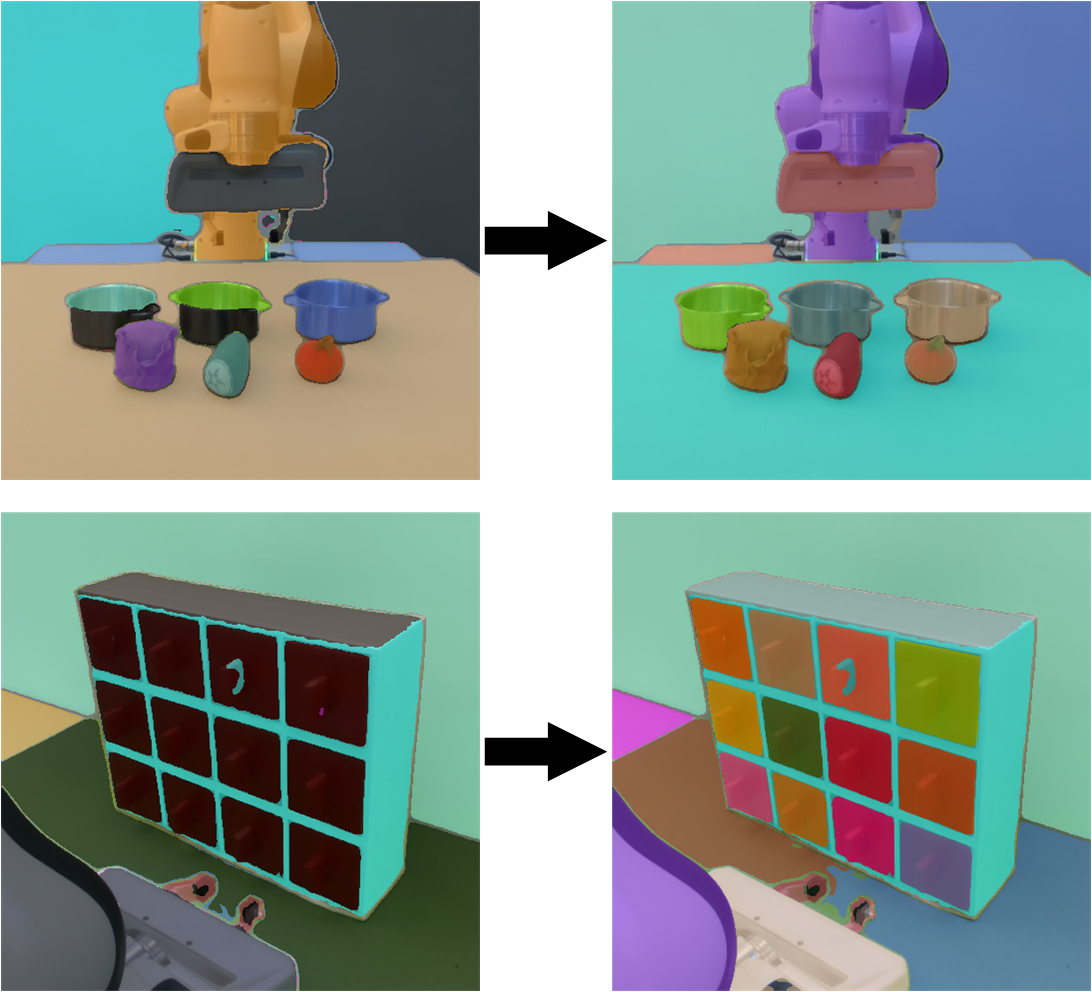}
\caption{Filtering to mitigate overlapping masks and combine related masks.} 
\label{fig:filtering}
\end{figure}

For better quality masks, the masks generated by Semantic-SAM are filtered based on patches and overlap, operating according to the logic outlined in Algorithms \ref{alg:patch} and \ref{alg:overlap} respectively. After these filtering steps, masks with an area lower than a threshold are removed. See Figure \ref{fig:filtering} for examples of the effects of filtering. The mask patch filter separates non-connected patches of a mask into separate masks, while retaining the holes in the original masks. The mask overlap filter discards, combines, or subtracts masks based on their overlap.

\begin{algorithm}[ht]
\DontPrintSemicolon
\caption{Mask Patch Filter}\label{alg:patch}
\SetKwInOut{Input}{Input}
\SetKwInOut{Return}{Return}
% % Define small-size comments
\SetKwComment{Comment}{$\triangleright$\ }{}
\newcommand\mycommfont[1]{\footnotesize\ttfamily\textcolor{black}{\hspace*{-0.5em}#1}}
\SetCommentSty{mycommfont}
\Input{Input masks $\mathbf{M_{in}}$, output masks $\mathbf{M}=\{\}$}
 \For{$m$ $\mathrm{in}$ $\mathbf{M_{in}}$ }{
    $\mathbf{P}= m.\mathrm{patches}()$ \\
    \For{$p$ $\mathrm{in}$ $\mathbf{P}$}{
         \eIf{$\neg \mathrm{has\_parent}$($p$)}{ 
            \tcp*[l]{The patch is not a hole}
            $\mathbf{M}$.append($p$)
         }
         {
            \tcp*[l]{Subtract the hole from the parent mask}
            $q$ = $p$.parent \\
            $q$  = $q \land \neg p$
         }
    }
}
\Return{$\mathbf{M}$}
\end{algorithm}

\setlength{\algomargin}{0pt}
\begin{algorithm}[ht]
\DontPrintSemicolon
\caption{Mask Overlap Filter}\label{alg:overlap}
\SetKwInOut{Input}{Input}
\SetKwInOut{Return}{Return}
% % Define small-size comments
\SetKwComment{Comment}{$\triangleright$\ }{}
\newcommand\mycommfont[1]{\footnotesize\ttfamily\textcolor{black}{\hspace*{-0.5em}#1}}
\SetCommentSty{mycommfont}

\Input{Input masks $\mathbf{M}$, upper threshold $u$, lower threshold $l$} 
$\mathbf{A} \gets \mathbf{M}.\mathrm{areas()}$ \tcp*[r]{Mask areas}
$\mathbf{I} \gets \mathrm{argsort}(\mathbf{A}, \text{descending})$ \tcp*[r]{Sorted indices}
$\mathbf{K} \gets [\ ]$ \tcp*[r]{Indices of masks to keep}

\While{$\mathbf{I} \neq \emptyset$}{
    $i \gets \mathbf{I}.\mathrm{pop()}$ \\
     \tcp*[l]{Get total overlap between with currently processed mask and all other masks}
    $c \gets \bigvee_{j \neq i} \mathbf{M}[j]$ \tcp*[r]{Coverage}
    $o_t \gets \frac{(c \wedge \mathbf{M}[i]).\mathrm{sum()}}{\mathbf{M}[i].\mathrm{sum()}}$ \tcp*[r]{Total overlap}    

    \If{$o_t < u$}{
        $s \gets \mathrm{zeros\_like}(\mathbf{M}[i])$ \tcp*[r]{Pixels to subtract}
        $J \gets \mathbf{I}.\mathrm{copy()}$ \\
        \While{$J \neq \emptyset$}{
            $j \gets J.\mathrm{pop()}$ \\
            \tcp*[l]{Get pairwise overlap for the currently processed mask pair ($i$,$j$)
            } 
            $o_p \gets \frac{(\mathbf{M}[i] \wedge \mathbf{M}[j]).\mathrm{sum()}}{\mathbf{M}[i].\mathrm{sum()}}$ \tcp*[r]{Pairwise overlap}

            \eIf{$o_p > l$}{
                $\mathbf{M}[i] \gets \mathbf{M}[i] \vee \mathbf{M}[j]$ \tcp*[r]{Combine}
                $\mathbf{I}.\mathrm{remove}(j)$ 
            }{
                $s \gets s \vee \mathbf{M}[j]$ 
            }
        }
        $\mathbf{M}[i] \gets \mathbf{M}[i] \wedge \neg s$ \tcp*[r]{Subtract}
        $\mathbf{K}.\mathrm{append}(i)$ 
    }
}
\Return{$\mathbf{M}[\mathbf{K}]$}
\end{algorithm}

Semantic-SAM can generate masks at different granularity, this parameter is adjusted for each test environment, as described in Section \ref{sec:experiments}. While \acp{vla} typically process low-resolution images (\textit{e.g.,} 224*244 for OpenVLA \cite{kim2024_openvla}), the \ac{methodname} framework enables processing at a higher resolution in the \ac{vlm} (we use 480x480), which further improves its chances of selecting the correct object instances. 

To highlight the object instances relevant to the task, we apply a relatively dense mask on all other pixels in the image using a semitransparent gray mask with alpha value 0.8. 
The relatively slow and expensive \ac{vlm} processes only the first image in a manipulation sequence, and the masks of relevant object instances are propagated to following images using a SAM2 \cite{ravi2024sam2} implementation for real-time video streams \cite{Zheng2024realtimeSAM2}.

We also study a variant of the framework, \ac{methodname}-relabeled, where the language instruction given to the \ac{vla} is simplified, while the \ac{vlm} gets the original instruction. The rationale for giving the original instruction to the \ac{vla} in \ac{methodname} without relabeling, is that the model may learn to balance the importance of the highlighting mask (as masking or selection via \ac{vlm} may fail) and the language instruction while generating actions. One potential downside with this approach is that the \ac{vla} may need to trust the highlighting mask more in unseen settings, but it would not have learned to do so from the training data. Conversely, if the language instruction given to the \ac{vla} is simplified, this reduces potentially conflicting information between the highlighting mask and the \ac{vla}'s interpretation of the language instruction, and more clearly offloads the language understanding responsibility to the \ac{vlm}. The downside of this approach is that when the highlighting fails due to problems in mask generation or mistakes in the \ac{vlm} during mask selection, the \ac{vla} will only get the incorrect data as input. We evaluate both approaches in Section \ref{sec:experiments}.

We run the process both on the training data to generate augmented training images, and during inference to highlight relevant objects. Ideally, the \ac{vla} learns to interact with the object instances deemed relevant by the \ac{vlm} which is equipped with superior semantic understanding capabilities.

\section{Experiments}
\label{sec:experiments}
The experiments were designed to investigate the following research questions:
\begin{enumerate}
    \item How well does a non-augmented \ac{vla} perform in tasks involving duplicate objects when subjected to concepts seen as well as unseen during demonstrations?
    \item Can the proposed method \ac{methodname} improve the \ac{vla} performances in the aforementioned scenarios?
    \item Does \ac{methodname} achieve better performance when provided with the original language instruction, or when given a simplified instruction so that task representation is carried primarily by the augmented image input?
\end{enumerate}

The experiments cover three task settings: lifting Lego Quatro blocks, putting vegetables into pots in a toy-kitchen setting, and opening drawers. For each task setting, three categories of instructions are defined:
\begin{itemize}
     \item \textbf{Category 1} contains language instructions that are seen in the training dataset. Except for the drawer setting, which consistently contains the same chest of drawers, we specify scene configurations by arranging objects. In these cases, each language instruction is assigned to a configuration that guarantees an unseen combination during testing.
     \item \textbf{Category 2} contains language instructions that are unseen combinations of concepts seen during training.
     \item \textbf{Category 3} contains language instructions that require extrapolation of concepts seen during training.
\end{itemize}

OpenVLA \cite{kim2024_openvla} is used as a baseline, and compared against \ac{methodname} with OpenVLA as the \ac{vla} and GPT-4.1 \cite{openai2025gpt4-1} as the \ac{vlm}. Joint models are trained for the blocks and kitchen tasks using their combined data, because they are both tabletop manipulation tasks that share a common camera viewpoint. The drawers task has a separate camera viewpoint and initial robot configuration, so separate models are trained for this task. To answer the third research question, a second pair of augmented models are trained, for which we replace the language instructions during training and inference with \textit{"lift the highlighted block"}, \textit{"put the highlighted vegetable in the highlighted pot"} and \textit{"open the highlighted drawer"}. To generate the augmented images during training and inference, we use Semantic-SAM granularity levels 1-3 or levels 1-4, mask overlap filter $u=0.8, l=0.4$, and mask area filter threshold 600 or 400 pixels, for the tabletop and drawers settings, respectively.

The models were finetuned with the OpenVLA default values: batch size 16, learning rate 0.0005, and LoRA \cite{hu2021loralowrankadaptationlarge} rank 32 and dropout 0.0. The image augmentation provided by OpenVLA is enabled during all training. The models for tabletop manipulation tasks are trained for 25 000 steps, and the models for the drawer opening task are trained for 50 000 steps as we used a larger dataset due to the precise movement needed when opening drawers.

\subsection{Lifting blocks}
\label{sec:lifitng_blocks}

\begin{figure}[b]
\centering
    \includegraphics[width=0.4\textwidth]{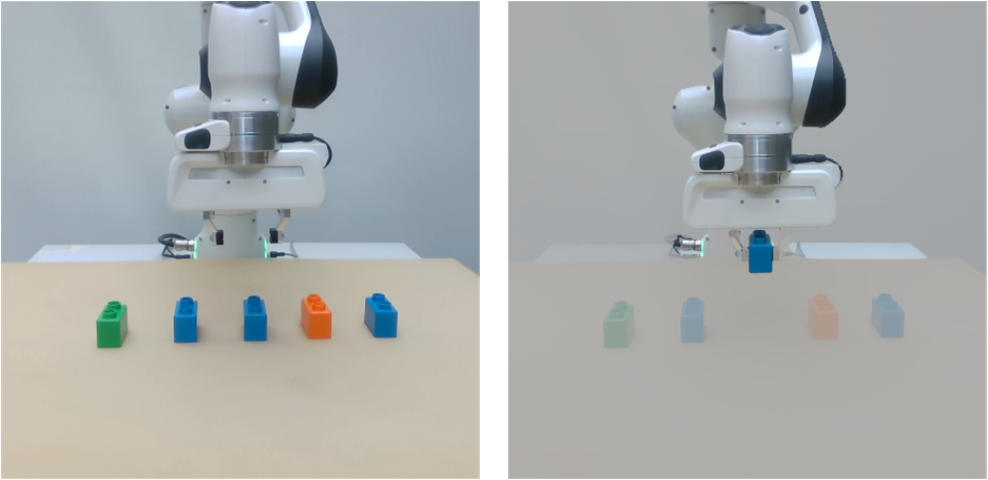}
    \setlength{\belowcaptionskip}{-6pt}
    \caption{Example of the task \textit{"lift the second blue block from the right"}. The unprocessed initial image is on the left, and the augmented image with propagated masks at the end of the task is at the right.} 
    \label{fig:blocks}
\end{figure}

\textbf{Task definition.} 
The blocks setting (see Figure \ref{fig:blocks}) is designed so that the required motion is simple, lifting a block roughly aligned with the initial gripper orientation from a row with a varying number of identically sized blocks, but successful task completion requires complex semantic and visual understanding to recognize the target block.
The studied instructions have the structure \textit{"lift the \{leftmost\,/\,rightmost\} \{orange\,/\,green\,/\,blue\} block"} and \textit{"lift the \{second\,/\,third\,/\,fourth/\,fifth\} \{orange\,/\,green\,/\,blue\} block from the \{left\,/\,right\}"}. 
We gather 120 demonstrations across 12 language instructions covering a subset of the instructions above. In this setting the scene configuration is defined by the number, color, and order of the blocks. The semantic concepts are the colors and positions of the blocks. Therefore, Category 2 tasks combine seen block positions with unseen block colors for those positions, and Category 3 tasks contain unseen ways to refer to block positions.

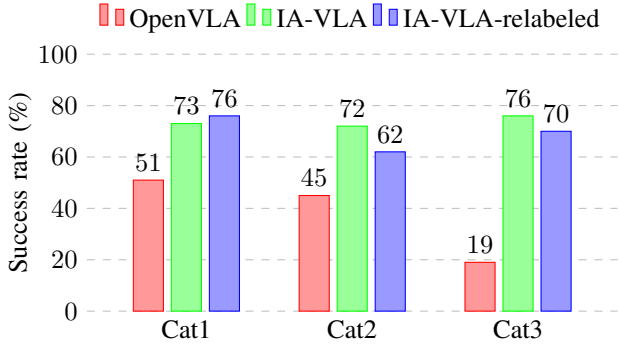
\begin{figure}[t]
    \vspace{1em}
    \centering
    \begin{tikzpicture}
    \begin{axis}[
        ybar, ymin=0, ymax=100,
    	symbolic x coords={Cat1,Cat2,Cat3},
        xtick=data,
        enlarge x limits=0.30,
        ylabel=Success rate (\%),
        ybar=3pt,
        bar width = 0.4cm,
    	legend style={draw=none,at={(0.5,1.22)},
    	anchor=north,legend columns=-1},
        nodes near coords,
        ymajorgrids=true,
        grid style=dashed,
        height=5cm,
        width=\columnwidth,
        axis line style={draw=none},
        ytick style={draw=none},
        xtick style={draw=none},
        xticklabel style={yshift=1ex},
        ylabel style={yshift=-1ex},
    ]
    \addplot[draw=red, fill=red!40!white] 
    	coordinates {(Cat1,51) (Cat2,45)
    		 (Cat3,19)};
    \addplot[draw=green, fill=green!40!white]
    	coordinates {(Cat1,73) (Cat2, 72)
    		 (Cat3,76)};
    \addplot[draw=blue, fill=blue!40!white]
    	coordinates {(Cat1,76) (Cat2, 62)
    		 (Cat3,70)};
    \legend{OpenVLA, \ac{methodname}, \ac{methodname}-relabeled}
    \end{axis}
    \end{tikzpicture}
    \caption{Results for lifting blocks.}
    \label{fig:blocks_result}
\end{figure}

\textbf{Results.} 
%For each test case in Table \ref{tab:block_matrix} we carry out five 
For each combination of block color and position, five evaluation runs are performed for each of the three models investigated, for a total of 450 evaluation runs. For each run, the model is given 30 seconds to perform the task, excluding the pre-processing time for the augmented models. A full point is given if the robot successfully lifts the block, and a half point is given if the robot attempts to grasp the correct object but fails to complete the task. All per-category results are reported as the percentage of maximum possible points in Figure \ref{fig:blocks_result}.

The results show that using either augmented model drastically increases success rates in all task categories, with the largest benefit naturally obtained in Category 3 instructions, which are most dissimilar to the training data. Interestingly, a large performance increase is obtained using the augmented approaches even in Category 1 tasks where the language instructions are familiar from the training data. This can be explained by the large variety of unseen configurations achievable by ordering up to six blocks in three colors, which makes understanding unseen configurations difficult even if the instruction is familiar. Therefore, a very nuanced semantic understanding is required to solve such tasks, which the large \ac{vlm} possesses but the finetuned \ac{vla} was unable to obtain. Note that the motion required to lift a block which is roughly aligned with the initial gripper orientation is very simple, so 120 demonstrations should be sufficient for a \ac{vla} to learn if it were not for the semantically complex instruction that specifies the block to lift.

\subsection{Filling pots}
\label{sec:filling_pots}

\begin{figure}[t]
\vspace{1em}
\centering
    \includegraphics[width=0.4\textwidth]{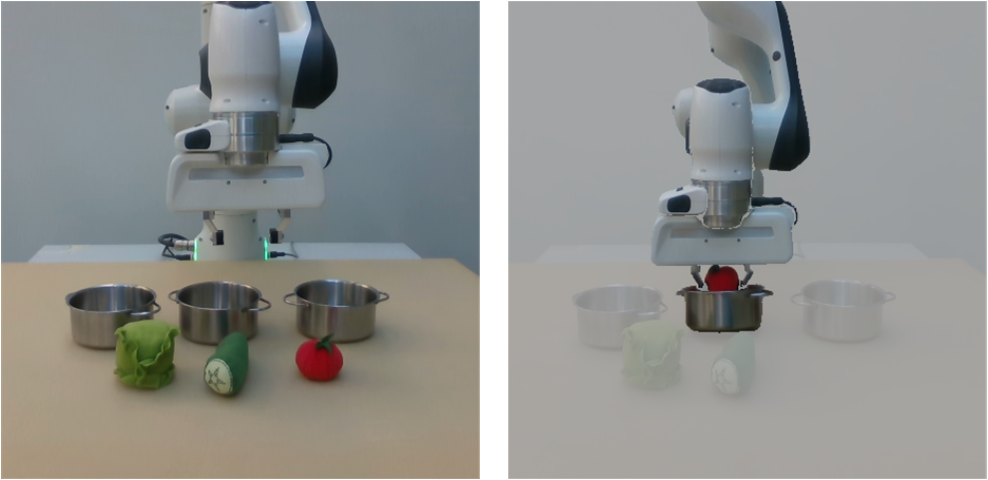}
    \setlength{\belowcaptionskip}{-6pt}
    \caption{Example of the task \textit{"put the tomato in the second pot from the left"}. The unprocessed initial image is on the left, and the augmented image with propagated masks at the end of the task is at the right.} 
    \label{fig:pots}
\end{figure}

\textbf{Task definition.} 
The toy kitchen setting (see Figure \ref{fig:pots}) involves a two-step pick-and-place motion, which is slightly more difficult than the lifting task of the blocks setting and more realistic in practical applications. 
The task is to put a uniquely identifiable vegetable in a duplicate pot, with two to four visually indistinguishable pots placed in a row. The semantic difficulty is lower than in the blocks setting, because there are fewer pots and there are no distractor containers. The studied instructions have the structure \textit{"put the \{tomato\,/\,cabbage\,/\,carrots\,/\,cucumber\} in the \{leftmost\,/\,rightmost\} pot"} and \textit{"put the \{tomato\,/\,cabbage\,/\,carrots\,/\,cucumber\} in the \{second\,/\,third\,/\,fourth\} pot from the \{left\,/\,right\}"}. We gather 120 demonstrations across 12 language instructions, covering a subset of the instructions above. In this setting the scene configuration
is defined by the number of pots together with the selection and order of vegetables. The semantic concepts are the names of the vegetables and the positions of the pots. Therefore, Category 2 tasks combine seen pot positions with unseen vegetables for that position, and Category 3 tasks contain unseen ways to refer to pot positions. It is worth noting that only half of the cases in Category 3 are fully unseen, while the other half contains descriptions of locations which have been seen in the blocks environment data, which the model is also trained on.

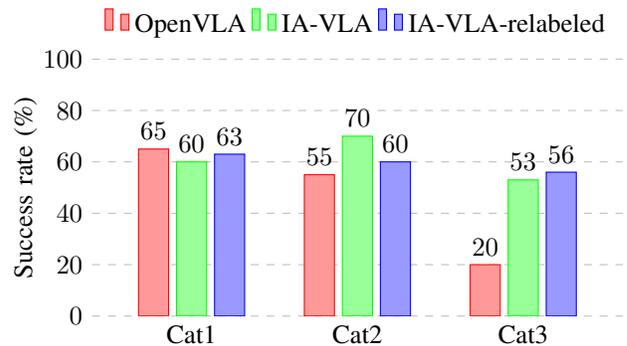
\begin{figure}[b]
    \centering
    \begin{tikzpicture}
    \begin{axis}[
        ybar, ymin=0, ymax=100,
    	symbolic x coords={Cat1,Cat2,Cat3},
        xtick=data,
        enlarge x limits=0.30,
        ylabel=Success rate (\%),
        ybar=3pt,
        bar width = 0.4cm,
    	legend style={draw=none,at={(0.5,1.22)},
    	anchor=north,legend columns=-1},
        nodes near coords,
        ymajorgrids=true,
        grid style=dashed,
        height=5cm,
        width=\columnwidth,
        axis line style={draw=none},
        ytick style={draw=none},
        xtick style={draw=none},
        xticklabel style={yshift=1ex},
        ylabel style={yshift=-1ex},
    ]
    \addplot[draw=red, fill=red!40!white] 
    	coordinates {(Cat1,65) (Cat2,55)
    		 (Cat3,20)};
    \addplot[draw=green, fill=green!40!white] 
    	coordinates {(Cat1,60) (Cat2, 70)
    		 (Cat3,53)};
    \addplot[draw=blue, fill=blue!40!white] 
    	coordinates {(Cat1,63) (Cat2, 60)
    		 (Cat3,56)};
    \legend{OpenVLA, \ac{methodname}, \ac{methodname}-relabeled}
    \end{axis}
    \end{tikzpicture}
    \setlength{\belowcaptionskip}{-6pt}
    \caption{Results for placing toy vegetables in pots.}
    \label{fig:pots_result}
\end{figure}

\textbf{Results.} 
For each combination of vegetable and pot position, five evaluation runs are performed for each of the three models investigated, for a total of 480 evaluation runs. For each run, the model is given 30 seconds to perform the task, excluding the pre-processing time for the augmented models. A full point is given if the robot successfully puts the vegetable in the pot, and a half point is given if the robot has successfully grasped the vegetable and navigated to the correct pot, but fails to insert the vegetable. All per-category results are reported as the  percentage of maximum possible points in Figure \ref{fig:pots_result}.

For Category 1 tasks, the baseline model gets the highest success rate, with both augmented models scoring almost as well. Also, the baseline model does not perform dramatically worse on Category 2 tasks compared to the augmented models. This is in contrast to what was observed in the block-lifting setting, and can be explained with the reduced variation in the range of possible configurations that can be achieved. The vegetables are always uniquely identifiable, and the 120 training demonstrations contain plenty of data to learn to identify them from, so the semantic complexity comes from understanding the relative positions of the pots. The variety of possible positions with four pots at most is significantly less than what can be achieved with the up to six blocks of the block lifting setting, where the interleaved blocks of different color caused additional visual complexity. The low performance of the baseline \ac{vla} in Category 3 tasks shows that even when the range of variation is smaller, the \ac{vla} still struggles to extrapolate from seen concepts.

\subsection{Opening drawers}
\label{sec:opening_drawers}

\textbf{Task definition.} 
The drawers setting represents tasks which require both complex semantic understanding, as the instruction now indicates the position in a two-dimensional grid rather than a one-dimensional row as in the previous settings, and precise motions.
The studied instructions have the structure \textit{"open the \{leftmost\,/\,rightmost\} drawer on the \{top\,/\,middle\,/\,bottom\} row"} and \textit{"open the \{second\,/\,third\,/\,fourth\} drawer from the \{left\,/\,right\} on the \{top\,/\,middle\,/\,bottom\} row"}.

Due to the difficulty of accurately grasping the drawer handle and pulling straight so that the drawer does not get stuck, fifty demonstrations for 12 language instructions, one referring to each drawer, are collected, resulting in 600 total demonstrations. There exists two ways to refer to each drawer (\textit{e.g.,} \textit{"third from left"} and \textit{"second from right"} refer to the same position on a row), so we are able to define both seen and unseen instructions. As the chest of drawers always looks the same and there are no other objects in the scene, there is no concept of a scene configuration in this setting, and all Category 1 tasks are fully seen in the training data. The semantic concepts refer to the order of the row and the position of a drawer in a row. Therefore, Category 2 tasks combine seen references to drawer positions with unseen rows for that position, and Category 3 tasks contain ways to refer to a drawer position which have not been shown for any row.

\begin{figure}[t]
\vspace{1em}
\centering
\includegraphics[width=0.4\textwidth]{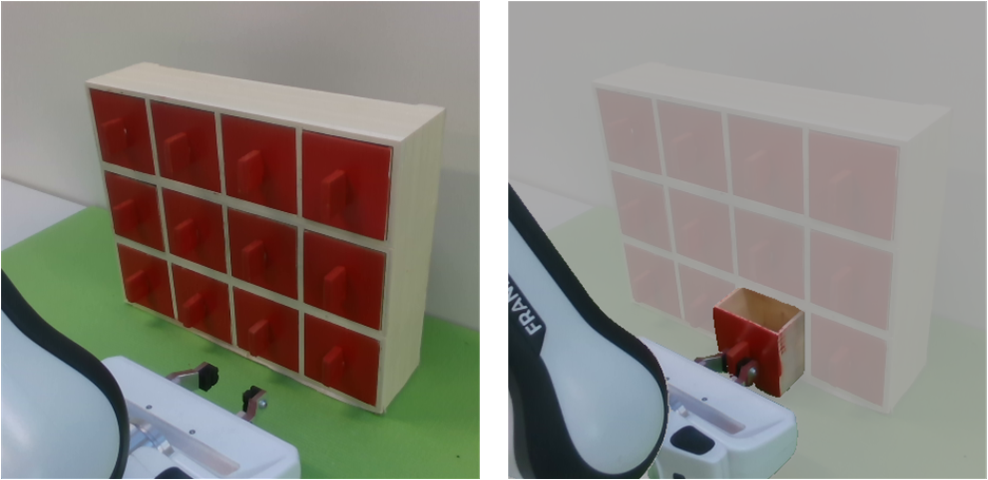}
\setlength{\belowcaptionskip}{-6pt}
\caption{Example of the task \textit{"open the third drawer from the left on the bottom row"}. The unprocessed initial image is on the left, and the augmented image with propagated masks at the end of the task is at the right.} 
\label{fig:drawers}
\end{figure}

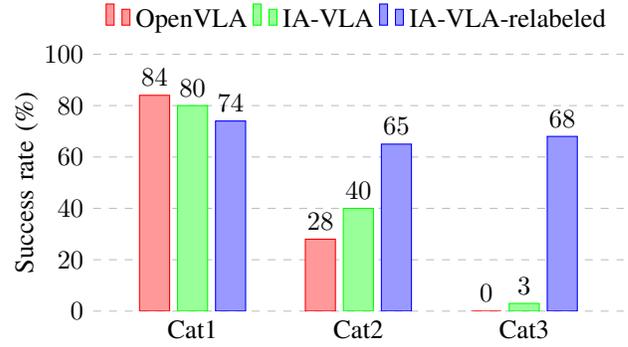
\begin{figure}
\vspace{1em}
    \centering
    \begin{tikzpicture}
    \begin{axis}[
        ybar, ymin=0, ymax=100,
    	symbolic x coords={Cat1,Cat2,Cat3},
        xtick=data,
        enlarge x limits=0.30,
        ylabel=Success rate (\%),
        ybar=3pt,
        bar width = 0.4cm,
    	legend style={draw=none,at={(0.5,1.22)},
    	anchor=north,legend columns=-1},
        nodes near coords,
        ymajorgrids=true,
        grid style=dashed,
        height=5cm,
        width=\columnwidth,
        axis line style={draw=none},
        ytick style={draw=none},
        xtick style={draw=none},
        xticklabel style={yshift=1ex},
        ylabel style={yshift=-1ex},
    ]
    \addplot[draw=red, fill=red!40!white]  
    	coordinates {(Cat1,84) (Cat2,28)
    		 (Cat3,0)};
    \addplot[draw=green, fill=green!40!white]  
    	coordinates {(Cat1,80) (Cat2, 40)
    		 (Cat3,3)};
    \addplot[draw=blue, fill=blue!40!white]  
    	coordinates {(Cat1,74) (Cat2, 65)
    		 (Cat3,68)};
    \legend{OpenVLA, \ac{methodname}, \ac{methodname}-relabeled}
    \end{axis}
    \end{tikzpicture}
    \setlength{\belowcaptionskip}{-6pt}
    \caption{Results for opening drawers.}
    \label{fig:drawers_result}
\end{figure}

\textbf{Results.} 
For each combination of row and drawer position, five evaluation runs are performed for each of the three models investigated, for a total of 360 evaluation runs. For each run, the model is given 30 seconds to perform the task, excluding the pre-processing time for the augmented models. A full point is given if the robot successfully opens the drawer, and a half point is given if the robot navigates to the correct drawer but fails to open it. All per-category results are reported as the percentage of maximum possible points in Figure \ref{fig:drawers_result}.

As the fixed size and appearance of the drawer does not allow for unseen configurations, Category 1 contained seen tasks. Therefore, it was expected that the baseline approach would perform well. Interestingly, the augmented approaches perform almost as well, even though the augmentation can introduce conflicting information at any stage, as\eg the mask generation or the mask selection with the \ac{vlm} the may fail.

Unlike in the tabletop settings, the augmented model without relabeling fails in Category 3. A larger difference in the success rate between the two augmented models is also observed in Category 2 tasks than in the tabletop tasks, with the model with original instructions performing significantly worse. In practice, we observe that the pre-processing usually highlights the correct drawer but the model moves to another drawer, indicating that it has not learned a successful balance of interpreting the language instruction and trusting the mask when the instruction contains unseen parts. 
We have a few theories for the reason for the degraded performance. First, the pretraining data for OpenVLA contains plenty of tabletop and toy-kitchen examples, so it more easily learns to operate in these environments. Second, even though the training dataset for the drawers was much larger than the one collected for the tabletop tasks, the range of tasks we can demonstrate is quite limited due to the fixed appearance of the chest of drawers. For example, demonstrations referring to the \textit{"second pot from the left"} in the toy kitchen environment can use anywhere from two to four pots, thus giving more context of what it means for an object to be in a certain location. In contrast, learning where the \textit{"second drawer from the left"} is, when the chest always has the same number of drawers per row, does not encourage the model to learn a higher-level understanding of the concept that would generalize to novel instructions. In contrast, the augmented model with relabeled instructions performs roughly equally well in all categories of a scene, which is expected because all tasks become equally seen when the instruction is simplified.

\subsection{Discussion and Failure modes}
The experiments gave insight into the computational overhead from using a pre-processing stage. Pre-processsing usually takes less than 10 seconds, with calling GPT taking the majority of the time and the other processing taking 2-3 seconds. Propagating the masks with SAM2 adds just 40 milliseconds of latency per \ac{vla} action.

In addition to the results specific to each task setting discussed in the previous section, some general trends can be observed. First, in all settings, the baseline model gets some points in Category 2 tasks, but less than the augmented models. This shows that the baseline \ac{vla} can sometimes combine concepts in the seen training data, but not consistently enough that it would not benefit from the highlighting the augmentation provides. In contrast, as the baseline model always gets worse scores in the unseen instructions of Category 3 this indicates that it has indeed learned to combine seen concepts in Category 2 sometimes, and it does not just get points by chance. Second, the baseline model consistently scores low on Category 3 tasks, showing that it is unable to extrapolate from concepts that have been shown in the training data. Here, the augmented models get greatly increased performance due to the superior reasoning of the larger \ac{vlm}. 

%205 total fails
%masking: 6 (2 original, 4 relabeled)
%VLM: 50 (17 original, 33 relabeled)
%execution: 143 (90 original 7blocks/43pots/40drawers, 53 relabeled 9blocks/36pots/8drawers)
%combined: 6 (6 original, 0 relabeled)
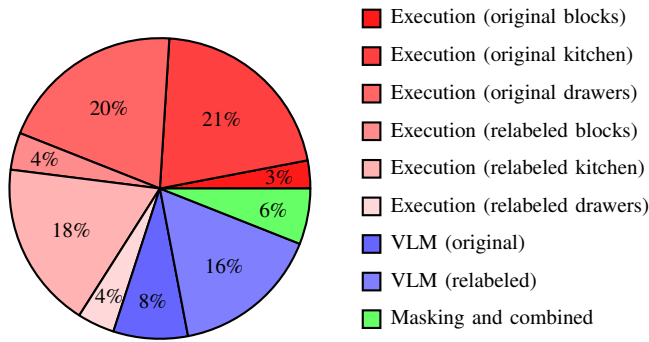
\begin{figure}[t]
    \vspace{1em}
    \centering
    \begin{tikzpicture}
    \tikzstyle{every node}=[font=\footnotesize]
    \pie [radius=2.0, text = legend, color = {red!90, red!75, red!60, red!45, red!30, red!15, blue!60, blue!50, green!60}]
        {3/Execution (original blocks), 21/Execution (original kitchen), 20/Execution (original drawers), 4/Execution (relabeled blocks), 18/Execution (relabeled kitchen), 4/Execution (relabeled drawers), 8/VLM (original), 16/VLM (relabeled), 6/Masking and combined}
    \end{tikzpicture}
    \caption{Failure modes of \ac{methodname}. Larger error categories separately show the failure percentages for the model with original instructions and the model with relabeled instructions, and the largest error category, execution errors, is further split between the different task settings.}
    \label{fig:failure_modes}
\end{figure}

\begin{figure}[t]
    \vspace{1em}
    \centering
    \begin{subfigure}[t]{0.47\columnwidth}
        \centering
        \includegraphics[width=\textwidth]{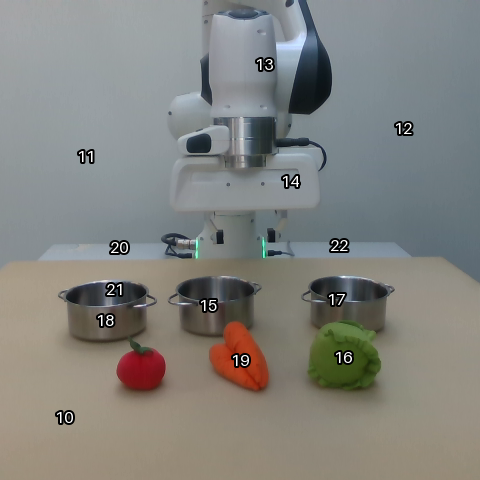}
        \caption{Due to a masking failure the tomato has not been assigned a mask, and thus there is no numeric tag indicating the tomato.}
    \end{subfigure}
    \hspace{0.02\columnwidth}
    \begin{subfigure}[t]{0.47\columnwidth}
        \centering
        \includegraphics[width=\textwidth]{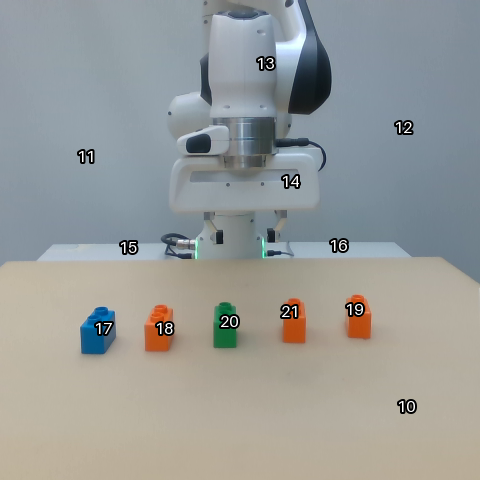}
        \caption{When presented with this image the \ac{vlm} erroneously reasoned \textit{"There are two orange blocks visible in the image, tagged 19 and 18. Counting from the right, the first orange block is 19, and the second orange block is 18."}.}
    \end{subfigure}
    \setlength{\belowcaptionskip}{-14pt}
    \caption{To the left there is an example of what the image annotated with numeric labels may look like when there has been a masking failure, and to the right there is an example of a scene with all necessary labels that the \ac{vlm} misinterprets when asked to "\textit{lift the second orange block from the right}".}
    \label{fig:mask_vla_failure_examples}
\end{figure}

We also study the reasons why tasks fail with \ac{methodname}.
The majority of failures, 70 \%, are failures where the \ac{vla} fails to carry out the task, even to such a degree to get partial points, although the correct objects has been masked. For example, such failures occur if the models misses the a vegetable when grasping and thus is not being able to move on to attempt placing it in the target pot. Another 24 \% of failures were occurred in cases where the relevant masks have been produced by Semantic-SAM and the filters, but the \ac{vlm} chooses the wrong numeric tags for any reason. A third type of failure occurred when the target object, or other objects needed to specify the relative position of the target object, had not been assigned numeric tags due to any failure in Semantic-SAM or the filtering we apply, but these accounted for only 3 \% of the failures. The remaining 3 \% of failures were cases where the wrong object is highlighted due to a failure during mask generation or selection with \ac{vlm}, and the \ac{vla} additionally fails to complete the task with the erroneously highlighted object. The error modes are summarized in Figure \ref{fig:failure_modes}, and in Figure \ref{fig:mask_vla_failure_examples} examples of a masking failure and a \ac{vlm} failure are provided.

Note that we only investigate test instances for failure reasons if the test ultimately fails, so cases where the masking or selection with \ac{vlm} fails but the execution still ends up being correct are not considered. This might explain why a somewhat higher fraction of the \ac{vlm} failure belong to the relabeled models, because when the original instruction is provided the \ac{vla} has a chance to use this information to successfully complete the task even when the \ac{vlm} has selected the wrong mask tags.

The failures due to mask generation were rare, and as \acp{vlm} improve we can expect the number of cases where it fails to select the numeric tags of the correct masks to decrease. Still, better mask quality would be desirable so that the relevant objects are more consistently fully highlighted with a single mask per object, and the background fully occluded. The current masking strategy may miss fine details, like the small fingers of the robot gripper, which is why we use a semitransparent mask.

The dominant failure mode, failures during \ac{vla} execution occurred mainly in the kitchen and drawers settings, but due to different reasons. Execution failures in the kitchen environment usually come from the robot missing the grasp on the target vegetable, and we expect these could be reduced with a larger and more varied demonstration set.
The execution errors in the drawers setting occurred almost exclusively with \ac{methodname} with original labels, because it failed consistently in Category 3 tasks and moved to different drawers than the ones indicated by the augmentation scheme, as discussed in Section \ref{sec:opening_drawers}.

Errors that resulted in failed evaluation runs, which we have analyzed in this section, occurred in just 205 out of the 860 total evaluation runs using augmented models in the three task settings. Some of these error types can be mitigated by augmenting the instruction in addition to the input image, and we expect the errors due to mask generation or selection to decrease as segmentation models and \acp{vlm} improve.

\section{Conclusions}
\label{sec:conclusions}
In this work, we presented \ac{methodname}, a framework that combines a high-performance \ac{vlm} with a fast \ac{vla} to tackle complex language instructions in robot manipulation tasks. Specifically, we investigated the problem of manipulating duplicate objects that has been underexplored in the \ac{vla} literature. Our results show that splitting the responsibilities of understanding complex language and creating real-time motion commands to separate models improves performance in tasks with complex language instructions that combine and extrapolate from concepts seen in the training data.
As segmentation models and \acp{vlm} improve, they can easily be used in our framework for increased performance.

A promising direction for future research include investigating whether, for \acp{vla} with multiple image inputs, augmenting only the main view suffices or whether consistent augmentation must be applied across all streams. Another direction would be to design a \ac{vla} model that can take highlighting masks as input in a separate channel, but this would require full retraining. Our framework could also be integrated with non-\ac{vla} methods that utilize segmentation for object selection, such as \cite{vosylius2025instantpolicyincontextimitation}. More broadly, the key challenge for \ac{vla} remains generalization to new tasks and environments, and while large training datasets are important in achieving this, further research is needed to find and address underexplored task types.

%%%%%%%%%%%%%%%%%%%%%%%%%%%%%%%%%%%%%%%%%%%%%%%%%%%%%%%%%%%%%%%%%%%%%%%%%%%%%%%%

\bibliographystyle{IEEEtran}
\bibliography{refs}

\end{document}